# ToLeRating UR-STD

Jan Feyereisl and Uwe Aickelin
School of Computer Science
The University of Nottingham
Nottingham, UK
jqf@cs.nott.ac.uk, uxa@cs.nott.ac.uk

## Abstract

*A new emerging paradigm of Uncertain Risk of Suspicion, Threat and Danger, observed across the field of information security, is described. Based on this paradigm a novel approach to anomaly detection is presented. Our approach is based on a simple yet powerful analogy from the innate part of the human immune system, the Toll-Like Receptors. We argue that such receptors incorporated as part of an anomaly detector enhance the detector's ability to distinguish normal and anomalous behaviour. In addition we propose that Toll-Like Receptors enable the classification of detected anomalies based on the types of attacks that perpetrate the anomalous behaviour. Classification of such type is either missing in existing literature or is not fit for the purpose of reducing the burden of an administrator of an intrusion detection system. For our model to work, we propose the creation of a taxonomy of the digital Acytota, based on which our receptors are created.*

## 1 The State of Affairs

Signature and anomaly based intrusion detection systems (IDS) have been around for many years, yet repeatedly have encountered the same pitfalls over and over again. This issue has been raised in the past, notably by Gates and Taylor [4] who challenged the idea of network based anomaly detection systems. Gates' and Taylor's account of the problem focuses mainly on anomaly detection in the network environment, however the area of host based intrusion detection, be it signature or anomaly based, exhibit analogous issues. Systems which combine these two approaches are sparse, let alone successful at attracting the security community to pursue the mixed approach further. We argue that such approach has so far been unsuccessful due to the way the fusion of the approaches was tackled.

One of the most counter-productive failings of existing IDS systems is the enduring presence of false positive alerts. This problem is aggravated even more with the sheer volume of such alerts. In anomaly based detection systems this is often due to the lack of re-training and re-calibration of detectors especially in unstable environments. These high volumes result in the difficulty of analysing logs produced by intrusion detection systems. Various forms of visualisation exist that attempt to solve this issue, however with little success [11]. Gates and Taylor [4] propose a classification of detected anomalies based on the type of attack that caused them in order for an IDS administrator to be able to prioritise and deal with most serious threats first. We propose that the incorporation of our immunological analogy enables anomaly classification at a level suitable for such prioritisation.

A seemingly simple part of the innate immune system that only recently began its research journey by immunologists has been our vital inspiration. This is the area of Pathogen Associated Molecular Patterns (PAMP) and Toll Like Receptors (TLR) in particular [3]. After some study of the security field, the authors realised that numerous researchers have presented ideas that support our proposed model, or that add functionality to form a more coherent whole outlining a paradigm worth investigating. We call this paradigm UR-STD.

## 2 The Uncertain Risk of STD

In 1994 Polly Matzinger shook the world of immunology with her controversial model of how the immune system operates [7]. Her view divided the immunological community which until this day segregates itself into classical immunology, the **Danger** model and immunologists who don't classify themselves into any of the aforementioned camps. In Matzinger's model, danger is defined as follows [7]:

- **Danger** - *the total damage to cells indicated by distress signals that are sent out when cells die an unnatural death*

The authors cannot determine if the shift in immunology from a concrete deterministic concept of self/non-self to a less concrete danger model had an effect on the thinking within the field of computer science, however a brief scan through research into new paradigms reveals a number of related approaches to computer security. Hollebeek's paper [5] on the role of suspicion in model based IDS gives a fascinating account of two new concepts of **suspicion** and **uncertainty**. These are introduced as part of a forensic analysis model that employs such new ways of thinking about what could constitute malicious behaviour in post-analysis. His definitions are as follows [5]:

- **Suspicion** - *likelihood that a given event or pattern of events is evidence of malicious behaviour*

- **Uncertainty** - *likelihood that deductions are correct and the likelihood the observed behaviour is normal **under the assumption that no malicious behaviour is present***

A number of points from Hollebeek's work are complementing the thesis of this paper. Firstly the fact that in forensic analysis an analyst only needs a few clues in order to recognise an intrusion. This approach works well from a forensics point of view rather than anomaly detection, however as will be shown in Section 6.2, this does not have to be the case. Secondly a concept which is paramount to the Danger model by Matzinger, the notion of context, as being a vital part of an IDS. Hollebeek does not associate his views with immunology or the Danger model. The assumption therefore might be that an investigation of Matzinger's model might provide some additional inspiration, which can teach us something applicable to IDS.

Besides the already mentioned new metrics, Saydjari proposed a new measure for system security, **Risk** [13]. Another term, originally classified more as a subjective measure, nevertheless presented by Saydjari with all the qualities of a good metric. His definition is as follows [13]:

- **Risk** - *risk of an event happening is the probability of that event happening multiplied by the consequences (damage or loss) from that event*

- **Overall Risk** - *the sum of the risk of all the bad events(failures) that can be induced by a malicious attacker on an information system*

Last but not least the term **threat**, which has been used in information security for decades, only now beginning to stand as a valid measure in its own right. Sahinoglu for example defines threat in his model in the following way [12]:

- **Threat** - *the probability of the exploitation of some vulnerability or weakness within a specific time frame.*

We propose that a new paradigm is emerging in existing literature. This paradigm moves away from traditional forms of measures from quantifiably objective to seemingly more subjective ones. We call this the **U**ncertain **R**isk of **S**uspicion, **T**hreat and **D**anger paradigm.

In the remainder of our paper we will present a new approach to intrusion detection closely tied to the newly defined UR-STD paradigm, particularly to the notion of *danger*, based on Matzinger's model, *suspicion* by Hollebeek and a more general notion of *threat*.

## 3 PAMPing up Anomaly Detection

As proposed by Matzinger's model and many subsequent research works undertaken by immunologists, pathogen associated molecular patterns and their detectors play a vital role in detecting danger in the human body. One such type of PAMP detectors are Toll-Like Receptors which are described in more detail in Section 4. This paper proposes that PAMP receptors are a necessary extension to current anomaly detection techniques in order to obtain results from our systems that go beyond what has been achieved in both anomaly and signature based detection fields. Hollebeek [5] confirms this view, arguing that in forensics, an analyst does not analyse data in a naive way. In other words he cannot analyse data without knowing about what he is looking for. We suggest that receptors, such as TLRs are missing. As John McHugh stated in his work on locality, the understanding of our systems with which we work is necessary to the degree, which will allow us to identify necessary parts of malicious activities [9]. Surprisingly McHugh's intentions portray the functionality of TLRs in the human immune system very well.

### 3.1 Signatures and Anomalies

Gates' provocative discussion on the anomaly detection paradigm in the networking environment [4] points out a number of disparities between Denning's seminal paper on anomaly detection [2] and the way her model is used by many security researchers. Anomaly based detection is not performing as we would like and thus we could assume either that its concept is applied incorrectly, it is missing something or both. Anomaly detection is missing that little extra knowledge that forensic analysts have in mind when looking for clues in data logs or when Agatha Christie's famous Belgian detective Poirot is solving a murder mystery. This missing knowledge is what Hollebeek wonderfully termed as '*smelling the rat*' without a '*smoking gun*' [5]. In our proposed model anomalous behaviour is not malicious. We suggest that malicious behaviour exhibits certain features, analogous to biological malignant objects, that it cannot occur without, much like viruses/bacteria. This is especially true when diversity is a factor in terms of observed attributes of a system. Such unique features are the extra knowledge that we need in order to distinguish between anomalous and malicious behaviour. TLRs are the proposed way of storing and sensing such features when they occur. In order to obtain the necessary information about such features we propose the creation of a taxonomy of the digital Acytota as described in Section 5.

## 4 Toll-Like Receptors

Paramount to our model is the notion of lower level, low volume receptors, which detect by-products of malicious activity. Such receptors exist in the human body in the form of Toll-Like Receptors. In the following section we will introduce these structures and draw an analogy to computer security.

### 4.1 Function

Toll-like receptors are a set of receptors on the surface of immune cells, which act as sensors to foreign microbial products, more broadly described as PAMPs. Up to this date around twelve TLRs have been discovered in the human body, each of them sensing a specific protein or a set of proteins discharged by viruses and bacteria. These microbial products are of essential nature to the existence and function of the foreign entities, thus making it impossible for the bacteria and viruses to adapt and evade these receptors. A simple definition of TLRs is that they are the initial line of defence against pathogens attacking a system. They sound an alarm when they encounter certain virus or bacteria specific chemicals, which trigger a cascade of events potentially resulting in an immune response. One can think of these receptors as piano keys. A different sound is played when a different key or combination of keys is pressed at once. TLRs can be categorised based on their specialization. TLRs 1,2,4,5 and 6 mainly specialise in bacterial products, whereas TLRs 3,7,8 and 9 specialise in the detection of virus specific by-products and nucleic acids. This functionality of TLRs is a feature which is in many ways analogous to some problems computer security. The fact that different combinations of activated TLRs perform different actions make it possible for the idea to be used in a multi-dimensional environment such as intrusion detection.

### 4.2 Analogy

From the description of the functionality of TLRs, one can see that such receptors lend themselves to the missing component of anomaly detection. This is due to their simplicity yet powerful functionality. One can argue that in the human body TLRs detect ligands which are produced by parts of malignant entities which are not subject to mutation. This is very different from the nature of digital malicious entities which can mutate or morph between versions or even on their own (e.g. polymorphic virus). However we argue that due to the deterministic nature of machines the creation of a taxonomy of the digital Acytota should be possible, which at some level is representative of the malicious entities digital systems encounter. Such a taxonomy will have to evolve over time at some level, however this will be incomparable in terms of frequency and volume to the level at which current signature based systems are updated.

To enforce our argument for TLRs as the correct candidate for our analogy, we look back at Hollebeek's work. In his paper, an overall suspicion in his model is described as *'a function of the degree to which each resource, event, or inference is suspicious and the number of independent reasons we have for being suspicious of it'* [5]. Here we can see a clear parallel with the functionality of TLRs. Given that each resource, event and possibly an inference holds a set of TLRs, we can obtain a measure of suspicion, which together with existing anomaly based systems can give us a more decisive view of what is happening.

## 5 Towards a Taxonomy of the Digital Acytota

In order to be able to employ the functionality of TLRs in intrusion detection systems, we need to develop a taxonomy of the digital malicious world. This is a pre-requisite, without which the proposed model becomes less effective. Such a taxonomy would not only allow us to implement the notion of TLRs fully, but would also allow us to classify detected malicious activities based on their type. One existing analysis of system calls in terms of their threat level and class of possible malicious activity has been proposed by Bernaschi in [1]. He classified system calls into four different threat levels, based on what kind of attack they are most likely to be used for, and nine groups of system functionality. For example Bernaschi classified system calls *open*, *mount* and *link*, among others, as belonging to threat level group 1, which represents system calls generally used for attacks that allow for the full control of a system.

Examination of various other components, levels and abstractions within computer systems is desirable in order to obtain a more complete picture of the TLR like features that are exhibited by malicious entities. We can start by examining existing works of research, followed by our own analysis of the digital Acytota. In current literature we can find work which contains some useful information from this perspective. For example McHugh's [9] work on locality contains description of features that are present in locality information unique to certain types of malicious network activities. McHugh uses the aggressive network behaviour of a Code Red or SQL/Slammer as an example, which corresponds to our desired, seemingly unique features. Another candidate is the set of system call arguments. These exhibit numerous features which can be considered unique to attacks. We can also clearly define what types of system call arguments can present a higher risk to a system. For example an argument which contains the reference to the *passwd* file under the Unix OS can be potentially much more hazardous than a reference to a word processing document in a user's home directory. The inclusion of system call arguments in our TLR model should also ensure the detection of mimicry attacks.

## 6 Proof of Concept

In order to evaluate our model, we have developed a system which incorporates the afore-described model. This system was constructed by observing the human immune system and while maintaining some aspects of the biological functionality, the immune system was largely an inspiration rather than a guide. For the anomaly detector which incorporates TLRs, Kohonen's Self-Organising Maps (SOM) [6] have been used, due to their similarity to the role of tissue in the human body. Further justifications for the use of SOMs are described in the following section.

### 6.1 SOMs as a Somatic model

The idea of self-organising networks comes as an ideal analogy to an environment within which TLRs exist in the human body, tissue. This is based on their similar behaviour and functionality. Both self-organising networks and tissue have nodes/cells, which dynamically change based on the situation they are in. SOM networks adapt to data which is passed through them and in human tissue, cells die and new ones are created as the body needs them.

One such type of self-organising networks are the aforementioned Kohonen's self-organising maps (SOM) [6]. Kohonen's SOM is an algorithm which allows the input of highly multi-dimensional data which is reduced to a predefined smaller dimensionality in order to represent such complex data in a more understandable manner for both human and machine. It organises the input data within a low-dimensional map, based on the similarity between the incoming data items and items already present in the map. In this way, the algorithm automatically organises and puts similar data items in topologically close proximity of each other within the generated map.

This self-organisation presents a number of parallels between the digital and a biological system. Such as the facilitation of a possible localisation function based on the topological proximity of neighbouring nodes within a SOM [10], possibly of interest to the research on locality [9].

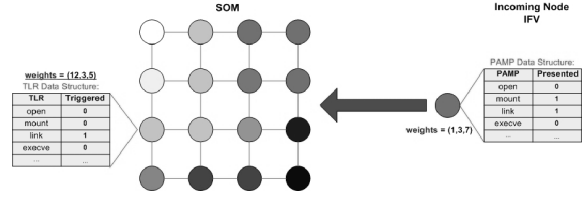

**Figure 1. TLR implementation within a SOM**

The ability of self-organising network algorithms such as Kohonen's SOM allow us to experiment and take into account the issues of diversity [8] with little difficulty. As will be shown in Section 6.3, the ability to monitor a large variety as well as only few aspects of the monitored system, provide an ideal environment for TLR implementation.

### 6.2 Implementation of TLRs

In our system, the implementation of TLRs is achieved by the creation of a table structure for every node within a SOM. We have taken Bernaschi's [1] system call threat level categorisation and experimented with threat level 1 group. This means system calls used in attacks which allow for the full control of a system. These 23 system calls are used in our model as one TLR, which each SOM node exhibits. Figure 1 shows the TLR model within the SOM context.

Kohonen's SOM algorithm, just like any other unsupervised learning algorithm, comprises of two steps. Training and detection. The role of TLRs in training is different from detection. The SOM, besides being trained on various system signals described below, learns the PAMPs encountered during training. In our implementations PAMPs are simply system calls which have been called during a time frame that is represented by an incoming node (data vector). The algorithm initially starts with random values assigned to the maps node's weight values, as proposed by Kohonen. The TLR data structure, as seen in Figure 1, is initialised for each node within the SOM map with a value for the *Triggered* field as *0*, meaning false. During training the map learns according to the standard SOM algorithm, with the additional functionality of activation of TLRs when relevant PAMPs are encountered. This activation occurs when incoming nodes on which the SOM is trained present PAMPs which the SOM node's TLRs bind to.

Incoming nodes into the SOM are our input feature vectors (IFVs). These comprise of the monitored features of the system in question. In our experimental setup, the IFVs constitute of 14 different process specific measures, such as CPU usage, memory usage and other host based metrics. For more information regarding the data used in our experiment see Section 6.3.1. As can be seen in Figure 1, PAMPs presented by IFVs are stored in a similar data structure as TLRs on nodes within the SOM. Once the SOM al-gorithm determines the winning node for the currently observed IFV, we simply trigger its TLRs depending on the presented PAMPs. This way we train our SOM not only on the IFV data but also on the specific features that are observable during normal operation of the monitored system/resource (i.e. a monitored process in our case). Once our SOM is trained, this represents a reduced representation of normal behaviour of the monitored system/resource.

In the detection stage, rather than adjusting the SOM map further, we simply find the Best Matching Unit (BMU) for the currently presented IFV and calculate their similarity. At the same time our system monitors if the currently observed IFV presents a PAMP. If it does, then the algorithm checks which TLRs of the BMU have been triggered during training. If the IFV presents PAMPs which were present during training, than we assume that their presence is justified, thus nothing additional happens. We might however retain the information holding the total number of triggered TLRs which were encountered during training for purposes yet to be investigated (e.g. SOM evolution). On the other hand if PAMPs are encountered which have not been detected during training, then our TLR activation level increases. This activation level is a measure of suspicious activity on its own, however we can use it to amend the result of the overall BMU distance measure, which is a representation of how anomalous a monitored IFV is. Thus making an IFV which presents a large number of PAMPs, not encountered during training, more anomalous and suspicious than an IFV which does not present any PAMPs. The effect of this action is to be further investigated and tested to fully appreciate its possibilities and correctness.

Using this model we argue that we are able to achieve what Hollebeek [5] proposed in his paper for forensics analysis. Yet we are looking to achieve this for real-time IDS. His point of view that an analyst only needs a few clues to recognise an intrusion becomes viable with the use of TLR's. Incompleteness as a necessary characteristic of the model is also pertinent in our case.

We further argue that the diversity of IFV's with the addition of TLR's offers a view of a context within which malicious activities are detected, which inevitably is a vital part of an IDS [5]. This leads onto another vital capability of our system, which has not been implemented as of yet, however it is expected to be realised in the foreseeable future. This is the ability to *classify* detected anomalies based on the type of attack that most likely caused them. In our model this is to be implemented by assigning individual TLRs to a set of features which, as defined in our taxonomy of Acytota, represent a certain class of attacks. For example a receptor called TLR1 will detect any PAMPs that are associated with denial of service attacks. For such attacks we already have a set of commonly used system calls as suggested by Bernaschi [1]. Other unique features that could belong to this receptor are for example specific types of IRC communication sequences exhibited by bots. Once our model contains a set of TLRs, each representing a class of attacks, we can incorporate various metrics from our UR-STD paradigm. For example we can define a measure of uncertainty, which depends on the proportion of activated TLRs and the total TLR activation level per SOM node. This will then provide us with a level of confidence showing to what degree a detected anomaly belongs to the chosen class of attacks. The choice of what class an anomaly belongs to is determined based on which TLR had the highest activation level. Additionally the calculation of risk of a particular type of anomaly, based on its classification is a possible candidate for further investigation.

The role of suspicion in our model is another metrics that can be used to make assumptions about the overall behaviour of the monitored system. For example we use it to denote the total number of IFVs which present at least one PAMP to the SOM which has not been encountered during training. Thus we can obtain a measure of how suspicious a session or a window of events is. This measure is described in Section 6.3.3.

### 6.3 Experimental Design

As mentioned previously, our experimental setup is mainly for illustrative purposes. It is yet to be statistically validated and further explored, however it provides a demonstration of our proposed TLR model with some initial results.

#### 6.3.1 Data

We have tested our model on a set of real data, based on the work of Twycross [14]. The data comprises of 55 sessions of normal behaviour and 5 attack sessions (4 sessions of a *wuftpd* vulnerability exploit and 1 *nmap* scan session). The session logs were generated using a process monitor, which samples various process specific information at a regular interval and system call information generated by the standard Unix tool called *strace*. A total number of 14 features have been observed using process monitor at any point in time, giving us a diverse coverage of the monitored system. For more information regarding the dataset please refer to [14].

#### 6.3.2 SOM parameters

A large number of parameters exists as part of the standard Kohonen's SOM algorithm. For our system we have chosen a set of parameters which have been statistically shown to be appropriate for a scenario such as our. These can be seen in Table 1. These parameters are a vital part of the anomaly detector and thus their thorough examination is a topic of future research.

**Table 1. SOM parameters**

| Parameter | Value |
|---|---|
| SOM Size | *15* |
| Iterations | *4* |
| Initial Learning Rate | *0.9* |
| Neighbourhood Function | *squared* |
| Neighbourhood Size | *15/2* |

One parameter which is used in our system that does not conform to the basic SOM algorithm is the use of a distance metrics other than Euclidean. A normalisation percentage metrics is used due to the fact that the data used in our experiments is not normalised at the pre-processing stage. The formula used for calculating this BMU % distance measure is shown below:

$$\frac{\sum Min(BMU, IFV)/(Max(BMU, IFV)/100)}{No. of Dimensions}$$

This measure gives us a clearer indication of the similarity between the tested IFVs and the trained SOM.

The system calls used as TLRs in our model are as follows: *open, link, unlink, chmod, lchown, rename, fchown, chown, mknod, mount, symlink, fchmod, execve, setgid, setreuid, setregid, setgroups, setfsuid, setfsgid, setresuid, setresgid, setuid, init_module*. As stated earlier they are threat level 1 group from Bernaschi's work [1].

The SOM was trained on 7 normal sessions, chosen at random from the existing 55 sessions. The sessions used for training are as follows: *2, 10, 14, 24, 35, 41, 57*. Following the training session, the SOM was tested against a subset of the normal sessions not used during training and against all attack sessions.

### 6.3.3 Results

Our initial experiments have produced the following promising results. In Figure 2 we can see the result of the BMU % distance measure for a subset of our tested normal sessions. From the graph we can see that the majority of IFVs belonging to the tested normal sessions have a percentage difference from the BMU in the range between 0 and 20%. Some false negatives in terms of BMU distance measure would be generated, depending on the chosen threshold for what percentage difference is classified as anomalous IFVs. Nevertheless if the threshold is set to 50%, then only a handful of IFVs are above such threshold and they belong to a small subset of tested sessions. It is correct to assume that by tweaking the SOM parameters and using a more precise BMU distance measure, these anomalous IFVs could be reduced even further.

Comparing Figure 2 to Figure 3, we can clearly see that the BMU % distance measure for the majority of tested

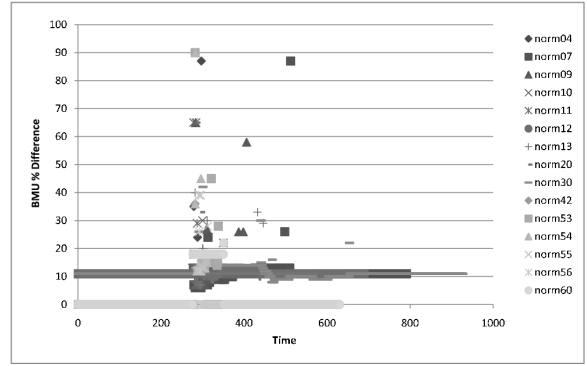

**Figure 2. BMU distance - normal sessions**

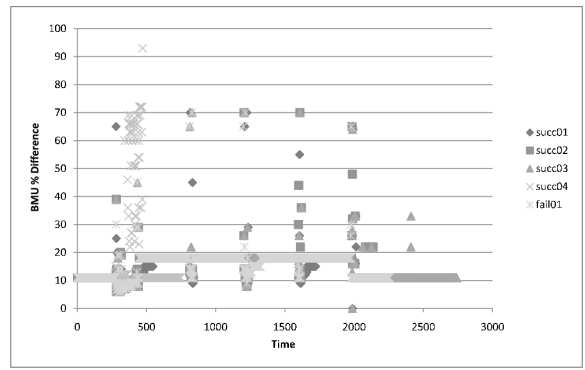

**Figure 3. BMU distance - attack sessions**

IFVs has on average increased. Also the number and spread of anomalous IFVs has increased and is present in all attack sessions. The session which produced the largest amounts of anomalous IFVs with a BMU distance of 50% and above is session 4 which is the *nmap* scan. This is possibly due to the fact that the *nmap* is a separate application from the one which is exploited by the wuftpd exploit. Thus session 4 produces more system calls and, as will be seen in Figure 4, a higher TLR activation level.

Figure 4 shows the levels of TLR activation across all tested sessions. It also shows the number of suspicious nodes (IFVs). TLR activation is the total number of TLRs which have been triggered during detection but not during training. An IFV can present up to 23 PAMPs, each of which, if presented, adds to the TLR activation level count. All attack sessions exhibit larger TLR activation levels in comparison to normal sessions. Session 4 exhibits a significantly higher TLR activation level, possibly due to the different composition of the underlying application. One could argue that the higher value of TLR activation for attack sessions is due to attack sessions containing larger amounts of IFVs. This might be the case and is currently under investigation, however it is assumed that using a larger diversity of TLRs, as proposed by the creation of a taxonomy of ma-licious entities, and with a corresponding dataset the TLR activation level results will be of more representative nature.

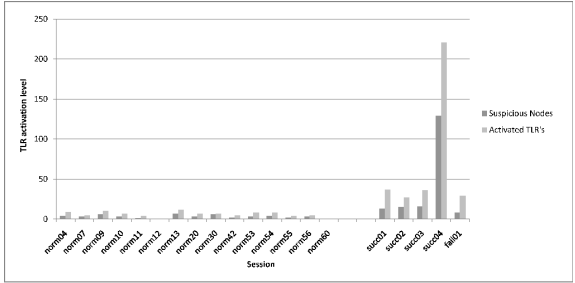

**Figure 4. TLR activation level**

One could argue that the difference between the BMU % distance of the normal and the attack sessions is not significant enough on its own, however when we look at the TLR activation levels as shown in Figure 4, we can see that there is a clear distinction between normal and attack sessions. We argue that further investigation of the SOM parameters will produce better results, at a level comparable or possibly higher than existing implementations of SOM based IDS systems. Once we correlate the results of BMU % distance with the TLR activation level we have a clear distinction between normal sessions and sessions that exhibit malicious behaviour.

## 7  Conclusion

The proposed Uncertain Risk of Suspicion, Threat and Danger paradigm points out to the information security field's tendency to increasingly incorporate, originally rather subjective, measures as part of security solutions. We suggest that the use of such measures can have a positive impact on the manageability of the outcome of intrusion detection systems, especially from an administrator's point of view. The incorporation of some of the proposed measures in the form of biologically inspired receptors as part of an existing anomaly detection technique shows promising results and points to the direction that such receptors might possibly be one of the missing links for successful anomaly detection in intrusion detection systems. Our view is that TLR's inclusion not only enhances the detection capability of anomaly detectors but also provides additional functionality by means of vital anomaly classification. The diverse nature of monitored features with the incorporated TLR functionality should also allow for the detection of, traditionally difficult to detect, attacks, such as race conditions and mimicry attacks. Further exploration of TLRs, their levels and their role as a classification mechanism is to be conducted in future research. The incorporation of further parts of the UR-STD paradigm are also to be investigated in order to exploit all the possibilities of the proposed paradigm.

## References

[1] M. Bernaschi, E. Gabrielli, and L. V. Mancini. REMUS: A security-enhanced operating system. *ACM Transactions on Information and System Security*, 5(1):36–61, Feb. 2002.

[2] D. Denning. An intrusion-detection model. *IEEE Transactions on Software Engineering*, 13(2):222–232, 1987.

[3] A. Dunne and L. A. J. O'Neill. The interleukin-1 receptor/toll-like receptor superfamily: signal transduction during inflammation and host defense. *Science's STKE 2003*, 2003:re3, 2003.

[4] C. Gates and C. Taylor. Challenging the Anomaly Detection Paradigm A provocative discussion. In *NSPW '06: Proceedings of the 2006 workshop on New security paradigms*, New York, NY, USA, 2006. ACM Press.

[5] T. Hollebeek and R. Waltzman. The role of suspicion in model-based intrusion detection. In *NSPW '04: Proceedings of the 2004 workshop on New security paradigms*, pages 87–94, New York, NY, USA, 2004. ACM Press.

[6] T. Kohonen. *Self-Organizing Maps*. Springer-Verlag, Berlin, 1996.

[7] P. Matzinger. Tolerance, danger, and the extended family. *Annual review of immunology*, 12(0732-0582):991–1045, 1994.

[8] R. A. Maxion. Use of diversity as a defense mechanism. In *NSPW '05: Proceedings of the 2005 workshop on New security paradigms*, pages 21–22, New York, NY, USA, 2005. ACM Press.

[9] J. McHugh and C. Gates. Locality: a new paradigm for thinking about normal behavior and outsider threat. In *NSPW '03: Proceedings of the 2003 workshop on New security paradigms*, pages 3–10, New York, NY, USA, 2003. ACM Press.

[10] M. Neal, J. Feyereisl, R. Rascuna, and X. Wang. Don't Touch Me, I'm Fine: Robot Autonomy Using an Artificial Innate Immune System. In *Proceedings of the 5th International Conference on Artificial Immune Systems (ICARIS 2006)*, number 4163 in Lecture Notes in Computer Science, pages 349–361, Oeiras, Portugal, 2006. Springer-Verlag.

[11] K. Nyarko, T. Capers, C. Scott, and K. Ladeji-Osias. Network intrusion visualization with NIVA, an intrusion detection visual analyzer with haptic integration. In *Haptic Interfaces for Virtual Environment and Teleoperator Systems, HAPTICS 2002*, pages 277–284, 2002.

[12] M. Sahinoglu. Security meter: a practical decision-tree model to quantify risk. *Security & Privacy Magazine, IEEE*, 3:18–24, May-June 2005.

[13] O. S. Saydjari. Risk: A good system security measure. In *COMPSAC '06: Proceedings of the 30th Annual In- ternational Computer Software and Applications Confer- ence (COMPSAC'06)*, pages 37–38, Washington, DC, USA, 2006. IEEE Computer Society.

[14] J. Twycross. *Integrated Innate and Adaptive Artificial Immune Systems applied to Process Anomaly Detection*. PhD thesis, The University of Nottingham, 2007.